\documentclass[runningheads]{llncs}
\usepackage{booktabs}
\usepackage{graphicx}
\usepackage[T1]{fontenc}
\usepackage{tabularx} 
\usepackage{amsfonts}
\usepackage{pifont} 
\usepackage{array}
\usepackage{multirow}
\usepackage{booktabs,tabularx}
\usepackage{xcolor}
\usepackage{graphicx}    
\usepackage{subcaption}   
\usepackage{caption}
\usepackage{amsmath}
\usepackage{graphicx}

\begin{document}
\title{Augment to Segment: Tackling Pixel-Level Imbalance in Wheat Disease and Pest Segmentation}
\titlerunning{Augment to Segment}
%
\author{Tianqi Wei\inst{1} \and
Xin Yu\inst{1} \and
Zhi Chen\inst{2} \and
Scott Chapman\inst{1} \and
Zi Huang\inst{1}
}
\authorrunning{T. Wei et al.}
%
\institute{The University of Queensland, Brisbane, Australia \\
\email{\{tianqi.wei,xin.yu,scott.chapman,helen.huang\}@uq.edu.au}\\
\vspace{0.1cm}
 \and
University of Southern Queensland, Toowoomba, Australia\\
\email{Zhi.Chen@unisq.edu.au}}
\maketitle

\begin{abstract}
Accurate segmentation of foliar diseases and insect damage in wheat is crucial for effective crop management and disease control. However, the insect damage typically occupies only a tiny fraction of annotated pixels. This extreme pixel-level imbalance poses a significant challenge to the segmentation performance, which can result in overfitting to common classes and insufficient learning of rare classes, thereby impairing overall performance.
In this paper, we propose a Random Projected Copy-and-Paste (RPCP) augmentation technique to address the pixel imbalance problem. 
Specifically, we extract rare insect-damage patches from annotated training images and apply random geometric transformations to simulate variations. The transformed patches are then pasted in appropriate regions while avoiding overlaps with lesions or existing damaged regions. In addition, we apply a random projection filter to the pasted regions, refining local features and ensuring a natural blend with the new background.
Experiments show that our method substantially improves segmentation performance on the insect damage class, while maintaining or even slightly enhancing accuracy on other categories. Our results highlight the effectiveness of targeted augmentation in mitigating extreme pixel imbalance, offering a straightforward yet effective solution for agricultural segmentation problems. 

\end{abstract}
\section{Introduction}

Wheat is one of the most widely cultivated crops and a key source of dietary calories worldwide~\cite{erenstein2022globalwheat,wang2025global}. However, its yield and grain quality are often threatened by a variety of diseases and insect pests, leading to substantial economic losses and posing serious challenges to global food security~\cite{fones2015impactSTB,savary2019globalburden}. Early and accurate detection of these threats is essential for effective crop protection, timely intervention, and sustainable management practices.
Recent advances in deep learning have enabled automated perception and analysis in agricultural vision tasks, providing an efficient and scalable alternative to traditional manual inspection~\cite{chen2024cf,ferentinos2018deepplantdetection,lim2024track,nigam2023deepwheat,thakur2022explainable,zheng2024pestyolo}.
Among various computer vision approaches, semantic segmentation has emerged as a powerful tool for pixel-wise recognition of disease and pest symptoms, enabling fine-grained characterization of lesion morphology and spatial distribution~\cite{polly2024semanticplants,shoaib2022deeptomatoseg,zhang2024pesttinysegformer}.

Despite its potential, applying semantic segmentation to wheat foliar disease datasets remains challenging due to large intra-class variation and severe class imbalance. Visual symptoms can differ significantly in size and appearance, complicating pixel-wise recognition. Rare classes, such as insect damage, often occupy only a tiny fraction of annotated pixels. This extreme pixel-level imbalance results in biased optimization, leading models to overfit on dominant classes and neglect rare classes. Even state-of-the-art models, such as SegFormer~\cite{xie2021segformer} achieve high accuracy on common classes like healthy tissue and STB lesions, yet their performance on insect damage regions remains substantially lower~\cite{zenkl2025towardsSTB}.

To address these challenges, we propose a targeted data augmentation strategy called Random Projected Copy-and-Paste (RPCP), which explicitly increases the representation of rare classes for training. RPCP first identifies insect-damage regions in annotated images and crops them according to their masks. The patches undergo geometric transformations, including random rotation and scaling, to generate variations. The transformed patches are then pasted into the appropriate regions of other training images while avoiding overlaps with existing damaged regions and lesions. After placement, a random projection filter refines local textures, blending the pasted regions naturally into the background and reducing visual artifacts. 
This straightforward yet effective pipeline mitigates extreme pixel-level imbalance by enriching the rare-class pixels. Moreover, it can integrate seamlessly into training workflows of any existing models without altering model architectures.

We conduct extensive experiments on a public wheat foliar disease segmentation dataset containing 3 classes: healthy leaves, Septoria tritici blotch (STB) lesions, and insect damage~\cite{zenkl2025towardsSTB}. The dataset exhibits a long-tailed pixel distribution, with insect damage accounting for only a small fraction of annotated pixels. To assess the generality of our approach, we evaluate multiple representative segmentation models, comparing each with and without RPCP. The results show that the RPCP augmentation consistently improves the performance of the rare class without degrading that of common classes.

To summarise, our main contributions are as follows:
\begin{itemize}
    \item We propose Random Projected Copy-and-Paste (RPCP), a rare-class-oriented augmentation strategy that mitigates extreme pixel-level imbalance in wheat foliar disease segmentation.
    \item RPCP is a model-agnostic strategy that can be seamlessly integrated into diverse existing training pipelines without requiring extra annotations or introducing architectural changes.
    \item Extensive evaluations on multiple representative models show that RPCP consistently improves rare-class accuracy without compromising common-class performance, highlighting its strong generalization capability.
\end{itemize}

\section{Related Work}
\noindent \paragraph{Plant Disease Semantic Segmentation.}
Plant disease recognition tasks include image-level identification~\cite{saad2024plant,ullah2023deepplantnet,wei2024plantwild,wei2024snap}, object localization~\cite{aldakheel2024detection,li2022improvedyolo,xie2020deepdetection}, and semantic segmentation~\cite{wei2024plantseg,yang2024ltdeeplav}. Among them, semantic segmentation provides the most detailed analysis by assigning pixel-level labels to diseased regions. Early segmentation methods mainly use convolutional neural network (CNN) encoder-decoder architectures, including CCNet~\cite{huang2019ccnet}, PSPNet~\cite{zhao2017pspnet}, and DeepLabV3 series~\cite{chen2017deeplabv3,chen2018deeplabv3+}. More recently, transformer-based architectures have been applied to plant disease segmentation, using global context modeling to capture long-range dependencies and complex lesion patterns~\cite{singh2024effective,xie2021segformer}. However, both CNN and Transformer-based methods often suffer from severe pixel-level class imbalance, which limits their ability to segment rare disease classes effectively.


\noindent \paragraph{Long Tail Learning.}
Long-tail distribution is a significant challenge in deep learning~\cite{yang2022longtailsurvey}, especially in natural scenes, where tail classes have fewer samples than head classes. This imbalance leads to biased learning, where models overfit to frequent classes and fail to generalize to rare ones. To alleviate this issue, existing methods can be broadly categorized into three groups. The first group is data-level strategies, such as re-sampling~\cite{gupta2019lvis,wu2020forest} and augmentation~\cite{chou2020remixaug,chu2020featureaug,zang2021fasa}. The second group includes model-level designs, including class-balanced losses~\cite{he2009learning,samuel2021distributionaloss,wang2021seesawloss}. Third, there are dedicated training paradigms like semi-supervised learning and domain adaptation~\cite{fan2022uccsemi,he2021redistributionsemi,hu2021semi}.

\noindent \paragraph{Data Augmentation for Semantic Segmentation.}
Data augmentation plays a crucial role in enhancing the robustness and generalization of segmentation models, particularly in conditions of limited or imbalanced data. There are widely adopted techniques such as flipping, scaling, and color jittering~\cite{shorten2019surveyaug}. More recent methods introduce region-level diversity by masking or blending patches~\cite{french2019semiperturbations}. Copy-paste augmentation, which pastes object-level regions into new contexts, has proven particularly effective for rare-class enhancement~\cite{fang2019instaboostcopy,ghiasi2021simplecp}. Based on this idea, our approach applies insect-specific pasting combined with a localized random projection to enhance the visual realism of augmented samples and improve robustness against distributional variations in segmentation performance.

\section{Method}
\begin{figure}[t]
    \centering
    \includegraphics[width=\linewidth]{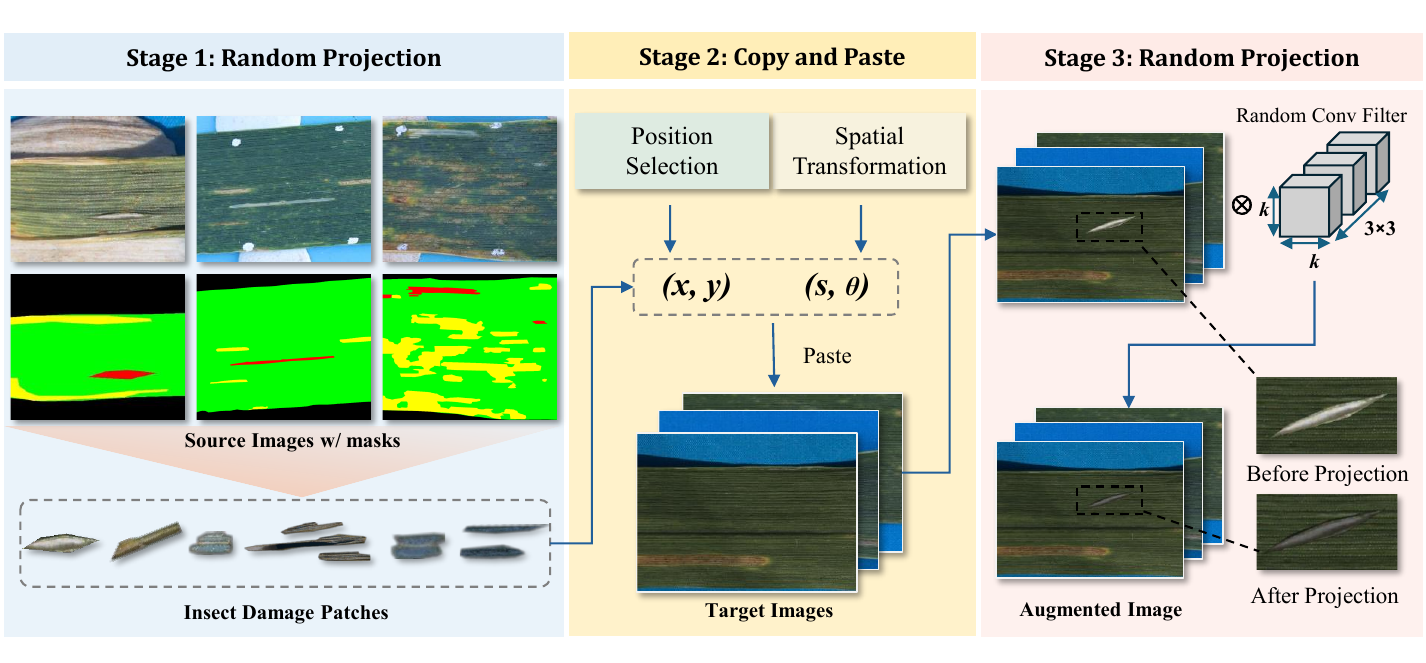}
    \caption{
    Overview of the proposed augmentation and refinement pipeline. 
    (1) \textbf{Rare-class Extraction}: insect-damage regions are cropped from annotated source images to construct patch candidates. 
    (2) \textbf{Insect Damage Copy-and-Paste}: patches are pasted into healthy regions of target images, with position selection \((x, y)\) and spatial transformations \((s, \theta)\) controlling placement, scale, and rotation. 
    (3) \textbf{Damage Pixel Random Projection}: a local random projection filter is applied to the pasted regions, enhancing texture blending and reducing artifacts to generate realistic augmented images.
}
    \label{fig:pipeline}
\end{figure}

In this section, we introduce our method for addressing the class imbalance in wheat leaf disease segmentation. The overview of task formulation and dataset characteristics is described in Sect.~3.1. The overall augmentation pipeline, including category-aware patch extraction and spatially constrained pasting, is presented in Sect.~3.2. The final component of our method, a region-wise random projection approach designed to regularize pasted areas and improve robustness, is detailed in Sect.~3.3. The overall framework is presented in Figure~\ref{fig:pipeline}. 

\subsection{Task Formulation}
Our study focuses on semantic segmentation for a specific type of wheat foliar disease, Septoria Tritici Blotch (STB), and insect damage.
Given an RGB image $I \in \mathbb{R}^{H \times W \times 3}$, the objective is to predict a per-pixel label map 
$Y \in \{0,1,\dots,C\}$, where each label corresponds to one of $C$ semantic classes (e.g., healthy leaf, disease lesions, and insect damage). 
This task exhibits a pronounced class imbalance. For example, insect damage regions are extremely scarce and characterized by irregular and localized patterns.
Such rare class pixels are typically small in size and provide limited training information, leading to suboptimal segmentation performance.
To mitigate this issue, we propose a Random Projected Copy-and-Paste (RPCP) augmentation strategy that enhances the representation of insect damage regions.

\subsection{Insect Damage Copy-and-Paste}
To address the significant class imbalance in wheat foliar disease segmentation, particularly the underrepresentation of regions with insect damage, we propose a targeted data augmentation pipeline. This pipeline enriches training images with additional rare-class instances through a two-stage process: (1) category-aware patch extraction and (2) spatially constrained pasting. 
In the first stage, annotated images are scanned to identify regions belonging to underrepresented classes, which are then cropped into patches with corresponding binary masks.
In the second stage, each patch is independently transformed by random rotation and scaling, 
and subsequently pasted into contextually appropriate locations. Candidate paste regions are restricted to valid areas (e.g., healthy leaf regions) while avoiding overlap 
with existing rare-class regions to preserve structural integrity. 
The pasting operation is parameterized by $(x,y,s,\theta)$, where $(x,y)$ specifies the placement position, 
$s$ controls the scaling factor, and $\theta$ denotes the rotation angle.
Formally, let $P \in \mathbb{R}^{h \times w \times 3}$ denote a transformed patch, 
$M \in \{0,1\}^{H \times W}$ its binary mask after transformation, and 
$I \in \mathbb{R}^{H \times W \times 3}$ the target image. 
The augmented image ${I'}$ and updated annotation $Y'$ are given by:
\begin{equation}
    I' = M \odot P + (1 - M) \odot I, \qquad 
    Y' = M \odot Y_P + (1 - M) \odot Y,
\end{equation}
where $Y_P$ is the patch label map and $\odot$ denotes element-wise multiplication. This process enhances the representation of rare-class instances while maintaining overall structural coherence.


\subsection{Damage Pixel Random Projection Refinement}
While class-aware augmentation directly improves class balance, the copy–paste operation may introduce artifacts and texture inconsistencies, potentially harming model generalization. To address these issues, we introduce a localized refinement method that operates only on pasted regions, altering their local appearance while preserving semantic structure.

Previous studies~\cite{vinh2016randomprojection,xu2021robustdistortion} have indicated that random convolution is an effective data augmentation approach, which can distort local textures while preserving the overall shape. Building on this property, we apply random projection to the pasted regions in our augmented images, enabling the creation of visually diverse appearances while maintaining label consistency. 
Given the augmented image ${I'} \in \mathbb{R}^{H \times W \times C}$ and its patch mask $Y_P \in \{0,1\}^{H \times W}$, we generate a random projection filter $\Theta \in \mathbb{R}^{h \times w \times C \times C}$. 
$H$, $W$ and $C$ represent the height, width and channels of ${I'}$, while $h$ and $w$ denote the height and width of $\Theta$. The weights of $\Theta$ are randomly sampled from a Gaussian distribution $\mathcal{N}(0, \sigma^2)$, with the hyperparameter $\sigma$ controlling the perturbation magnitude.
The perturbed image $X$ is obtained as:
\begin{equation}
X = I' * \Theta.
\end{equation}
We then blend $X$ with ${I'}$ inside the mask $Y_P$ using coefficient $\alpha$, producing the refined image $X'$:
\begin{equation}
X' = \alpha Y_P \odot X + (1 - \alpha Y_P) \odot {I'}.
\label{eq:blend}
\end{equation}

This refinement introduces spatially localized texture and color variations in pasted regions, mitigating unnatural boundaries and improving robustness.


\subsection{Training Objective}
To train the segmentation model $f_\theta$, we minimize a supervised loss on the augmented data. 
Specifically, we adopt the pixel-wise cross-entropy loss $\ell_{ce}$. 
Given a batch of augmented samples 
$\mathcal{B} = \{(X'_i, Y'_i)\}_{i=1}^{|\mathcal{B}|}$, 
the model generates predictions $\hat{Y}'_i = f_\theta(X'_i)$ for the $i$-th image. 
The supervised loss is then defined as
\begin{equation}
\mathcal{L}_{seg} = \frac{1}{|\mathcal{B}|} \sum_{i=1}^{|\mathcal{B}|} 
\frac{1}{H \times W} \sum_{j=1}^{H \times W} 
\ell_{ce}\big(\hat{Y}'_i(j),\, Y'_i(j)\big),
\end{equation}
where $H \times W$ denotes the spatial resolution and $j$ indexes a pixel in the image. 
Here, $\hat{Y}'_i(j) = f_\theta(X'_i)(j)$ represents the predicted probability vector at pixel $j$, 
and $Y'_i(j)$ is the corresponding ground-truth label. 

\section{Experiment}
\subsection{Setups}
\noindent \paragraph{Datasets.} The experiments are conducted on the  Septoria Tritici Blotch (STB) dataset introduced by Boulent et al.~\cite{zenkl2025towardsSTB}. STB is a high-resolution image dataset specifically curated for semantic segmentation tasks of wheat foliar disease and insect damage. It comprises 422 RGB images of wheat leaves captured under diverse lighting conditions, each with a resolution of 1024~$\times$~1024~px.
For the semantic segmentation task, each image is densely annotated with pixel-level masks for three semantic classes: healthy leaf area, necrotic tissue, and insect damage. Notably, the STB dataset exhibits a significant class imbalance, where healthy leaf regions dominate the pixel distribution, while insect damage appears sparsely. The skewed distribution introduces a strong bias toward majority classes, often resulting in poor performance on rare categories.


\noindent \paragraph{Evaluation Metrics.}
Intersection over Union (IoU) and pixel accuracy (Acc) are used to evaluate segmentation performance for each semantic class. IoU measures the overlap between the predicted and ground-truth regions, while Acc quantifies the proportion of correctly classified pixels. The metrics are defined as:
\begin{equation}
\text{IoU}_c = \frac{TP_c}{TP_c + FP_c + FN_c},
\qquad
\text{Acc}_c = \frac{TP_c}{TP_c + FN_c},
\end{equation}
where $TP_c$, $FP_c$, and $FN_c$ denote the numbers of true positive, false positive, and false negative pixels for class $c$. In addition, overall performance is reported using mean Intersection over Union (mIoU) and mean Accuracy (mAcc), computed by averaging the IoU and Acc values across all classes:

\begin{equation}
\text{mIoU} = \frac{1}{C}\sum_{c=1}^C \text{IoU}_c,
\qquad
\text{mAcc} = \frac{1}{C}\sum_{c=1}^C \text{Acc}_c,
\end{equation}
where $C$ denotes the total number of classes.

\noindent \paragraph{Baselines.}
We select a set of representative semantic segmentation models as the baselines, including both convolutional and transformer-based architectures. 
Specifically, we adopt representative CNN-based models with multi-scale context modules and encoder–decoder designs, such as PSPNet~\cite{zhao2017pspnet}, CCNet~\cite{huang2019ccnet}, and the DeepLabV3 series~\cite{chen2017deeplabv3,chen2018deeplabv3+}. For transformer-based approaches, we employ SegFormer~\cite{xie2021segformer} and SAN~\cite{xu2023sideadapterSAN}. 
In addition, we include advanced architectures such as SegNeXt~\cite{guo2022segnext} and ConvNeXt~\cite{liu2022convnext}, which combine the efficiency of CNN architectures with transformer-inspired contextual modeling to enhance segmentation performance.

\noindent \paragraph{Implementation.}
All models are trained using the MMSegmentation~\cite{mmseg2020} framework built upon PyTorch. 
We adopt the AdamW optimizer, with learning rates selected from the range 3e-5 to 1e-4 and weight decay values from \{0.0001, 0.0005, 0.001\}. Experiments are conducted using batch sizes $\{4, 8, 16\}$. 
During the training process, input images are first randomly resized within a scale ratio range of $[0.5, 2.0]$, then randomly cropped to $512 \times 512$, followed by horizontal flipping with a probability of 0.5 and color jittering to enhance robustness to illumination changes.

\newcommand{\diff}[1]{\textcolor{green!50!black}{\scriptsize{(#1)}}}
\newcommand{\diffneg}[1]{\textcolor{red!70!black}{\scriptsize{(#1)}}}
\begin{table*}[t]
\centering
{\footnotesize
\begin{tabularx}{\textwidth}{c|
>{\centering\arraybackslash}X >{\centering\arraybackslash}X|
>{\centering\arraybackslash}X >{\centering\arraybackslash}X|
>{\centering\arraybackslash}X >{\centering\arraybackslash}X|
>{\centering\arraybackslash}X >{\centering\arraybackslash}X}
\toprule
\multirow{2}{*}{\textbf{Method}} & \multicolumn{2}{c|}{\textbf{Class 1}} & 
\multicolumn{2}{c|}{\textbf{Class 2}} & 
\multicolumn{2}{c|}{\textbf{Class 3}} & 
\multicolumn{2}{c}{\textbf{Average}} \\
\cmidrule{2-9}
& \textbf{IoU} & \textbf{Acc} & \textbf{IoU} & \textbf{Acc} & \textbf{IoU} & \textbf{Acc} & \textbf{mIoU} & \textbf{mAcc} \\
\midrule
DeepLabV3 & 97.60 & 98.28 & 81.96 & 87.08 & 57.79 & 72.19 & 79.12 & 85.85 \\
w/ RPCP & 97.78 & 99.18 & 82.17 & 87.76 & 61.76 & 77.14 & 80.57 & 88.03 \\
$\Delta$ & \diff{+0.18} & \diff{+0.90} & \diff{+0.21} & \diff{+0.68} & \diff{+3.97} & \diff{+4.95} & \diff{+1.45} & \diff{+2.18} \\
\midrule
DeepLabV3+ & 97.29 & 98.45 & 81.93 & 88.29 & 58.89 & 75.61 & 79.37 & 87.45 \\
w/ RPCP & 97.75 & 99.03 & 84.47 & 89.85 & 62.32 & 78.99 & 81.51 & 89.29 \\
$\Delta$ & \diff{+0.46} & \diff{+0.58} & \diff{+2.54} & \diff{+1.56} & \diff{+3.43} & \diff{+3.38} & \diff{+2.14} & \diff{+1.84} \\
\midrule
PSPNet & 97.48 & 99.02 & 82.75 & 88.71 & 60.91 & 74.32 & 80.38 & 87.35 \\
w/ RPCP & 97.64 & 98.89 & 83.78 & 90.41 & 64.38 & 78.96 & 81.93 & 89.42 \\
$\Delta$ & \diff{+0.16} & \diffneg{-0.13} & \diff{+1.03} & \diff{+1.70} & \diff{+3.47} & \diff{+4.64} & \diff{+1.55} & \diff{+2.07} \\
\midrule
CCNet & 97.60 & 98.11 & 84.08 & 91.62 & 60.22 & 75.93 & 80.63 & 88.55 \\
w/ RPCP & 97.79 & 98.92 & 84.98 & 91.10 & 64.50 & 82.08 & 82.42 & 90.70 \\
$\Delta$ & \diff{+0.19} & \diff{+0.81} & \diff{+0.90} & \diffneg{-0.52} & \diff{+4.28} & \diff{+6.15} & \diff{+1.79} & \diff{+2.15} \\
\midrule
SAN & 97.53 & 98.85 & 83.14 & 90.10 & 62.73 & 76.18 & 81.13 & 88.38 \\
w/ RPCP & 97.57 & 98.90 & 83.20 & 90.17 & 65.86 & 77.32 & 82.21 & 88.80 \\
$\Delta$ & \diff{+0.04} & \diff{+0.05} & \diff{+0.06} & \diff{+0.07} & \diff{+3.13} & \diff{+1.14} & \diff{+1.08} & \diff{+0.42} \\
\midrule
SegFormer & 97.38 & 98.01 & 82.73 & 90.12 & 68.16 & 79.87 & 82.76 & 89.33 \\
w/ RPCP & 97.98 & 99.12 & 85.82 & 91.67 & 72.36 & 82.57 & 85.39 & 91.12 \\
$\Delta$ & \diff{+0.60} & \diff{+1.11} & \diff{+3.09} & \diff{+1.55} & \diff{+4.20} & \diff{+2.70} & \diff{+2.63} & \diff{+1.79} \\
\midrule
ConvNeXt & 98.00 & 98.85 & 85.18 & 91.57 & 70.67 & 84.31 & 84.62 & 91.58 \\
w/ RPCP & 98.00 & \textbf{99.43} & 85.46 & 89.09 & 74.30 & \textbf{84.58} & 85.92 & 91.03 \\
$\Delta$ & \diff{+0.00} & \diff{+0.58} & \diff{+0.28} & \diffneg{-2.48} & \diff{+3.63} & \diff{+0.27} & \diff{+1.30} & \diffneg{-0.55} \\
\midrule
SegNeXt & 97.93 & 99.06 & 85.32 & 91.45 & 72.81 & 81.56 & 85.35 & 90.69 \\
w/ RPCP & \textbf{98.01} & 98.97 & \textbf{86.04} & \textbf{92.87} & \textbf{75.62} & 84.40 & \textbf{86.56} & \textbf{92.08} \\
$\Delta$ & \diff{+0.08} & \diffneg{-0.09} & \diff{+0.72} & \diff{+1.42} & \diff{+2.81} & \diff{+2.84} & \diff{+1.21} & \diff{+1.39} \\
\bottomrule
\end{tabularx}
}
\vspace{0.2cm}
\caption{Overall model performance. Green numbers indicate improvement, red numbers indicate decrease. $\Delta$ rows show the change relative to the baseline.}
\label{tab:main_results}
\end{table*}

\subsection{Main Results}
Table~\ref{tab:main_results} presents the segmentation performance of all baseline models and their RPCP-enhanced variants. 
For Class~1 (healthy leaf), all methods achieve very high IoU and Acc values, with differences generally below 1\%, indicating that this class is well-represented and easy to segment. 
Class~2 (lesion region) exhibits moderate variation across models, while all IoU values remain above 81\%. In contrast, Class~3 (insect damage) shows the largest performance gaps, reflecting the difficulty of segmenting underrepresented features.  
Among the baselines, SegNeXt achieves the strongest Class~3 performance with 72.81\% IoU and 81.56\% Acc, followed closely by ConvNeXt (70.67\% IoU, 84.31\% Acc).

Introducing RPCP consistently improves Class~3 results across a wide range of backbones, highlighting the robustness of the proposed augmentation.  
The largest IoU gains are observed for CCNet (+4.28\%) and SegFormer (+4.20\%), while DeepLabV3+ (+3.43\%) and PSPNet (+3.47\%) also exhibit substantial improvements.  
Moreover, the results on common classes indicate that RPCP does not sacrifice performance where training data is already abundant. Overall, RPCP yields clear benefits in both mIoU and mAcc for most models. For example, SegFormer achieves an mIoU increase from 86.26\% to 88.89\% (+2.63\%) and an mAcc increase from 87.54\% to 89.33\% (+1.79\%).
Though we observe a slight decline in the overall performance of ConvNeXt, the general trend confirms that RPCP is effective in enhancing rare-class segmentation while maintaining or improving accuracy for dominant classes.

\subsection{Ablation Study}

\begin{table}[t]
\centering
\small
\setlength{\tabcolsep}{4pt}
\begin{tabular}{c c | cc | cc| cc| cc}
\toprule
\multicolumn{2}{c|}{\textbf{Modules}} & \multicolumn{2}{c|}{\textbf{Class 1}} & \multicolumn{2}{c|}{\textbf{Class 2}} & \multicolumn{2}{c|}{\textbf{Class 3}} & \multicolumn{2}{c}{\textbf{Average}}\\
\midrule
\textbf{CP} & \textbf{RP} & \textbf{IoU} & \textbf{Acc} & \textbf{IoU} & \textbf{Acc} & \textbf{IoU} & \textbf{Acc} & \textbf{mIoU} & \textbf{mAcc} \\
\midrule
\ding{55} & \ding{55} & 97.93 & \textbf{99.06} & 85.32 & 91.45 & 72.81 & 81.56 & 85.35 & 90.69 \\
\midrule
\ding{51} & \ding{55} & 97.96 & 98.90 & 85.91 & 92.39 & 73.02 & 83.63 & 85.63 & 91.64 \\
\midrule
\ding{51} & \ding{51}& \textbf{98.01} & 98.97 & \textbf{86.04} & \textbf{92.87} & \textbf{75.62} & \textbf{84.40} & \textbf{86.56} & \textbf{92.08} \\
\bottomrule
\end{tabular}
\vspace{0.2cm}
\caption{Ablation study of the copy-paste (CP) and random projection (RP) modules. Both improve the segmentation performance.}
\label{tab:ablation_rpcp}
\vspace{-0.5cm}
\end{table}

To evaluate the effectiveness of the proposed RPCP method, we conduct an ablation study by incrementally adding its two components: the class-oriented Copy-Paste (CP) module and the Random Projection (RP) module on top of SegNeXt. As shown in Table~\ref{tab:ablation_rpcp}, starting from the baseline without targeted augmentation, adding CP improves the IoU and accuracy of the rare Class 3 (insect damage) from 72.81\% to 73.02\% and from 81.56\% to 83.63\%, respectively, while maintaining similar performance on other classes. This demonstrates the effectiveness of CP, which directly increases the exposure of rare-class patterns during training and enables the model to learn more discriminative features for insect damage without disturbing the distribution of dominant classes.

Building on CP, incorporating the RP module leads to further improvements. The largest gains are observed on Class 3, where IoU increases to 75.62\% and accuracy to 84.40\%. 
This suggests that enhancing local texture blending and reducing artifacts in pasted regions results in more consistent predictions.
Overall, the combination of CP and RP achieves the highest performance, validating the effectiveness of both modules and showing that their complementary effects contribute to improved segmentation performance.

\subsection{Sensitivity Study of Hyperparameters}
\vspace{-0.2cm}
\noindent \paragraph{Number of Pasted Patches per Image.}
We investigate the impact of varying the number of pasted patches per image, denoted as $k$, on segmentation performance.During the copy-and-paste augmentation process, we set $k \in \{0,1,2,3,4\}$. 
As $k$ increases, the total pasted area naturally expands, enlarging the coverage of the rare insect-damage class.
We choose SegNeXt as the segmentation model. Each experiment is repeated 5 times and averaged to reduce randomness.
As shown in Table~\ref{tab:number_paste}, the rare-class performance of SegNeXt improves consistently when 1 or 2 patch regions are added, indicating the effectiveness of moderate augmentation. However, as the number continues to increase, a significant drop in accuracy occurs. These results highlight that while moderate copy-and-paste augmentation effectively boosts performance for rare classes, excessive augmentation may disrupt natural spatial context and create a distribution mismatch between the augmented training set and the test set, resulting in degraded performance.

\begin{table}[t]
\centering
\setlength{\tabcolsep}{4pt}
\begin{tabular}{c| c c |c c| c c| c c}
\toprule
\multirow{2}{*}{\textbf{Number}} &
\multicolumn{2}{c|}{\textbf{Class 1}} &
\multicolumn{2}{c|}{\textbf{Class 2}} &
\multicolumn{2}{c|}{\textbf{Class 3}} &
\multicolumn{2}{c}{\textbf{Average}} \\
\cmidrule(lr){2-9}
& \textbf{IoU} & \textbf{Acc} & \textbf{IoU} & \textbf{Acc} & \textbf{IoU} & \textbf{Acc} & \textbf{mIoU} & \textbf{mAcc} \\
\midrule
0 & 97.93 & \textbf{99.06} & 85.32 & 91.45 & 72.81 & 81.56 & 85.35 & 90.69 \\
1 & \textbf{98.01} & 98.97 & \textbf{86.04} & \textbf{92.87} & \textbf{75.62} & \textbf{84.40} & \textbf{86.56} & \textbf{92.08} \\
2 & 97.97 & 98.65 & 85.61 & 92.25 & 74.82 & 84.32 & 86.13 &  91.74\\
3 & 97.91 & 98.60 & 85.38 & 92.61 & 72.22 & 83.64 & 85.17 &  91.62\\
4 & 97.76 & 98.51 & 85.54 & 91.75 & 69.23 & 78.88 & 84.18 &  89.71\\
\bottomrule
\end{tabular}
\vspace{0.2cm}
\caption{Influence of the number of pasted patches per image, with the best performance achieved using a single patch.}
\label{tab:number_paste}
\vspace{-0.8cm}
\end{table}

\noindent \paragraph{Size of Random Projection Filter.}
We conduct experiments to evaluate the impact of varying the spatial size of the random projection filter $\Theta$. The size of $\Theta$, denoted as $(h, w)$, determines the spatial extent over which local texture variations are introduced. A small filter enables detailed texture adjustments while preserving sharp semantic boundaries, but may be insufficient for removing large-scale appearance discrepancies. In contrast, a larger filter can better integrate the pasted regions into the background but may over-smooth the fine-grained details and cause blurred boundaries. We fix all other hyper-parameters and vary $(h, w) \in \{3\times3, 5\times5, 7\times7, 9\times9\}$. The results shown in Table~\ref{tab:filter_size} indicate that the smallest filter size (3$\times$3) results in the best overall performance, indicating that fine-scale texture refinement enhances segmentation without oversmoothing. Larger filters gradually reduce accuracy, with 9$\times$9 causing a significant drop, which is likely due to excessive smoothing of insect-damage boundaries.

\begin{table}[t]
\centering
\setlength{\tabcolsep}{4pt}
\begin{tabular}{l| c c| c c| c c| c c}
\toprule
\multirow{2}{*}{\textbf{Size}} &
\multicolumn{2}{c|}{\textbf{Class 1}} &
\multicolumn{2}{c|}{\textbf{Class 2}} &
\multicolumn{2}{c|}{\textbf{Class 3}} &
\multicolumn{2}{c}{\textbf{Average}} \\
\cmidrule(lr){2-9}
& \textbf{IoU} & \textbf{Acc} & \textbf{IoU} & \textbf{Acc} & \textbf{IoU} & \textbf{Acc} & \textbf{mIoU} & \textbf{mAcc} \\
\midrule
$3{\times}3$  & \textbf{98.01} & \textbf{98.97} & \textbf{86.04}& \textbf{92.87} & \textbf{75.62} & \textbf{84.40} & \textbf{86.56} & \textbf{92.08} \\
$5{\times}5$ & 97.85 & 98.74 & 86.00 & 91.86 & 74.35 & 83.20 & 86.07 & 91.27 \\
$7{\times}7$ & 97.86 & 98.69 & 85.95 & 91.61 & 72.86 & 82.78 & 85.56 & 91.03 \\
$9{\times}9$ & 97.88 & 98.70 & 84.84 & 91.63 & 68.98 & 81.53 & 83.90 & 90.62 \\
\bottomrule
\end{tabular}
\vspace{0.2cm}
\caption{Influence of the random projection filter size, with the best performance achieved using a $3 \times 3$ filter.}
\label{tab:filter_size}
\vspace{-0.2cm}
\end{table}

\noindent \paragraph{Perturbation Magnitude.} 
To offer further insight into the effect of the Random Projection Filter, we also investigate the effect of the perturbation magnitude $\sigma$ in the random projection filter $\Theta \sim \mathcal{N}(0,\sigma^{2})$. $\sigma$ controls the amplitude of pixel-wise appearance changes applied to the pasted regions. 
Small values produce only subtle variations, whereas large values introduce stronger texture and color changes. Specifically, we vary $\sigma \in \{0.00, 0.05, 0.10, 0.20, 0.30\}$, where $\sigma{=}0$ corresponds to disabling refinement. As depicted in Table~\ref{tab:perturbation_magnitude}, $\sigma=0.20$ achieves the highest performance,  
suggesting that moderate perturbations best balance blending and detail preservation.

\begin{table}[t]
\centering

\setlength{\tabcolsep}{4pt}
\begin{tabular}{c| c c| c c| c c| c c}
\toprule
\multirow{2}{*}{\textbf{Magnitude}} &
\multicolumn{2}{c|}{\textbf{Class 1}} &
\multicolumn{2}{c|}{\textbf{Class 2}} &
\multicolumn{2}{c|}{\textbf{Class 3}} &
\multicolumn{2}{c}{\textbf{Average}} \\
\cmidrule(lr){2-9}
& \textbf{IoU} & \textbf{Acc} & \textbf{IoU} & \textbf{Acc} & \textbf{IoU} & \textbf{Acc} & \textbf{mIoU} & \textbf{mAcc} \\
\midrule
0.00& 97.93 & \textbf{99.06} & 85.32 & 91.45 & 72.81 & 81.56 & 85.35 & 90.69 \\
0.05& 97.91 & 98.88 & 85.51 & 92.73 & 73.11 & 82.08 & 85.51 &  91.23\\
0.10& 97.92 & 98.89 & 85.45 & 92.76 & 73.96 & 82.14 & 85.78 & 91.26 \\
0.20& \textbf{98.01} & 98.97 & \textbf{86.04}& \textbf{92.87} & \textbf{75.62} & \textbf{84.40} & \textbf{86.56} & \textbf{92.08} \\
0.30& 97.88 & 98.73 & 85.63 & 92.72 & 70.98 & 81.01 & 84.83 & 90.02 \\
\bottomrule
\end{tabular}
\vspace{0.2cm}
\caption{Influence of the perturbation magnitude $\sigma$. The best segmentation performance is observed at $\sigma=0.20$.}
\vspace{-0.7cm}
\label{tab:perturbation_magnitude}
\end{table}

\noindent \paragraph{Blending Coefficient.} 
We explore the effectiveness of the blending coefficient $\alpha$ in Equation~\ref{eq:blend}. 
We vary $\alpha \in \{0.2, 0.4, 0.6, 0.8, 1.0\}$.
As shown in Table \ref{tab:blending_coefficient}, performance generally improves with larger $\alpha$, indicating that stronger blending enhances the integration of pasted regions and reduces artifacts. The peak is observed at $\alpha = 0.8$, suggesting an optimal balance between preserving background continuity and incorporating augmented content. However, when $\alpha$ reaches $1.0$, performance drops, likely because excessive blending overrides natural background information and disrupts contextual consistency.

\begin{table}[t]
\centering
\setlength{\tabcolsep}{4pt}
\begin{tabular}{c| c c| c c| c c| c c}
\toprule
\multirow{2}{*}{\textbf{Coefficient}} &
\multicolumn{2}{c|}{\textbf{Class 1}} &
\multicolumn{2}{c|}{\textbf{Class 2}} &
\multicolumn{2}{c|}{\textbf{Class 3}} &
\multicolumn{2}{c}{\textbf{Average}} \\
\cmidrule(lr){2-9}
& \textbf{IoU} & \textbf{Acc} & \textbf{IoU} & \textbf{Acc} & \textbf{IoU} & \textbf{Acc} & \textbf{mIoU} & \textbf{mAcc} \\
\midrule
0.2& 97.81 & 98.96 & 85.33 & 91.56 & 72.84 & 83.35 & 85.33 & 91.29 \\
0.4&   97.94 & \textbf{98.99} & 85.56 & 92.01 & 72.96 & 83.58 & 85.49& 91.53 \\
0.6& 97.80 & 98.87 & 85.96 & 91.82 & 73.42 & 84.78 & 85.73 & 91.82 \\
0.8& \textbf{98.01} & 98.97 & \textbf{86.04}& \textbf{92.87} & \textbf{75.62} & {84.40} & \textbf{86.56} & \textbf{92.08} \\
1.0& 97.97 & 99.00 & 85.71 & 92.13 & 74.29 & \textbf{84.75} & 85.99 & 91.96 \\
 
\bottomrule
\end{tabular}
\vspace{0.2cm}
\caption{Influence of the blending coefficient $\alpha$, with the best overall performance observed at $\alpha=0.8$.}
\label{tab:blending_coefficient}
\vspace{-0.7cm}
\end{table}

\subsection{Analysis}
\vspace{-0.7cm}
\noindent \paragraph{Visualization Analysis.}

\begin{figure}[t]
    \centering
    \includegraphics[width=\linewidth]{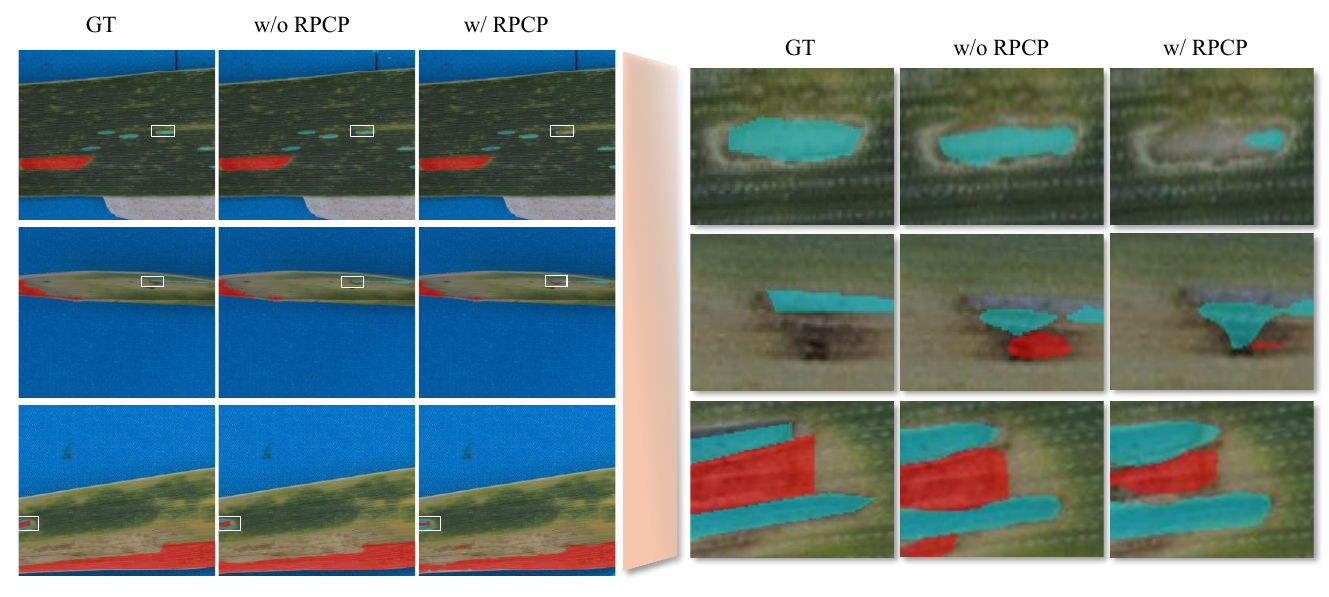}
    \caption{SegNeXt segmentation results with and without RPCP. Red masks denote leaf necrosis areas, and cyan masks denote insect damage regions.}
    \label{fig:rpcp_visual}
\end{figure}

Figure~\ref{fig:rpcp_visual} provides a qualitative visual comparison of SegNeXt predictions with and without RPCP. As shown in the figure, red parts indicate Class 2 (Leaf Necrosis), and cyan parts represent Class 3 (Insect Damage). The zoomed-in regions on the right reveal that, without RPCP, the model tends to underestimate insect-damage areas. Due to the class imbalance, some insect-damage regions are even misclassified as lesions. In contrast, with RPCP, such misclassification is alleviated and the segmentation of insect damage becomes more complete and coherent. This visualization also explains the slight performance gain observed for Class 2, since reducing false positives of insect damage leads to more precise leaf necrosis segmentation.

\begin{figure}[t]
    \centering
    \begin{subfigure}[b]{0.49\textwidth}
        \centering
        \includegraphics[width=\linewidth]{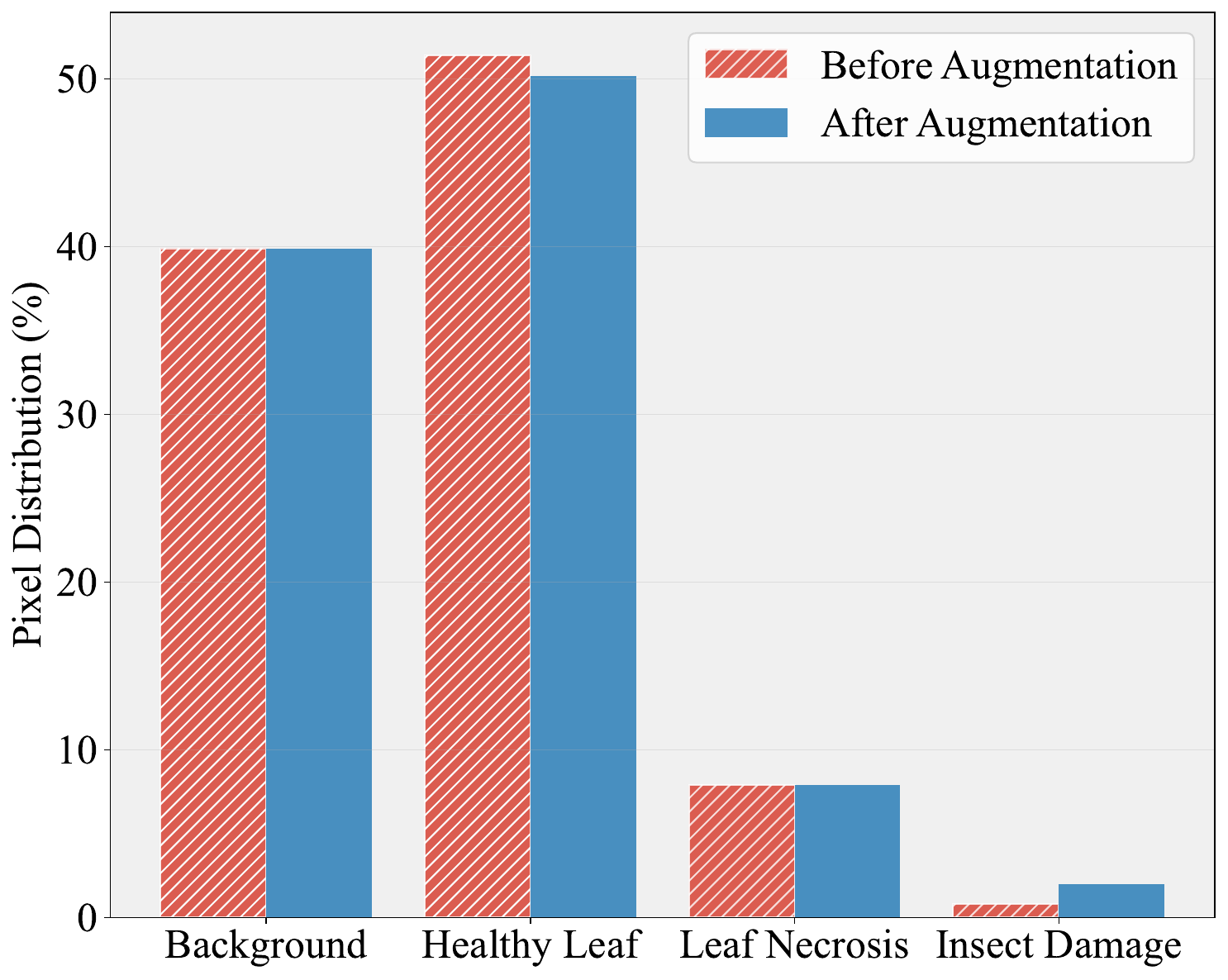}
        \caption{Pixel distribution.}
        \label{fig:pixel_dist}
    \end{subfigure}
    \begin{subfigure}[b]{0.49\textwidth}
        \centering
        \includegraphics[width=\linewidth]{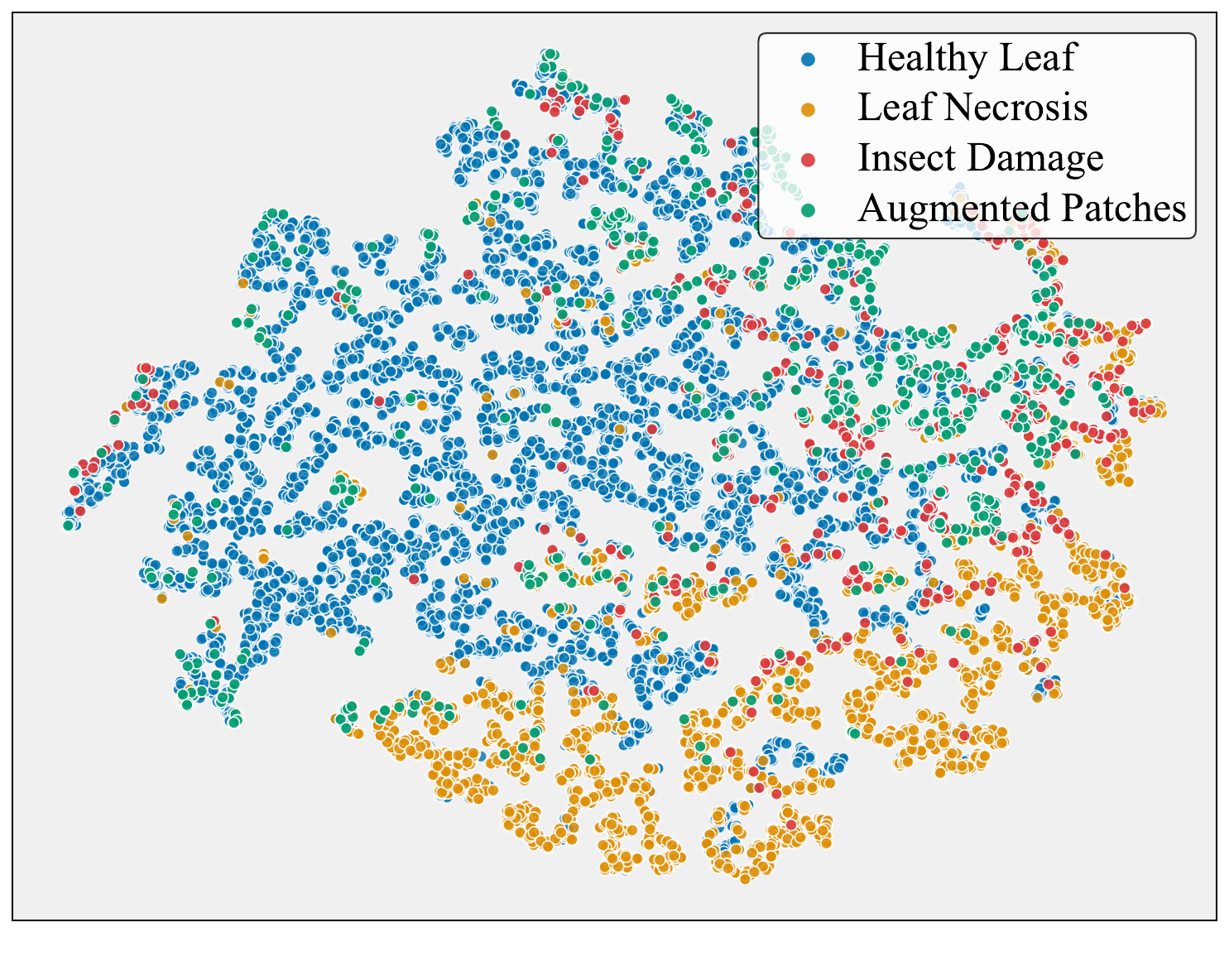}
        \caption{t-SNE visualization.}
        \label{fig:tsne_vis}
    \end{subfigure}
    \caption{
    (a) Pixel-level distribution shows the alleviation of class imbalance. 
    (b) The t-SNE visualization of sampled RGB pixels indicates that augmented patches share similar color distributions with authentic insect-damage regions.}
    \label{fig:rpcp_effect}
\end{figure}

\noindent \paragraph{Pixel Distribution after RPCP.} 
Figure~\ref{fig:rpcp_effect}(a) presents the pixel-level distribution across categories. After applying RPCP, the proportion of insect-damage pixels is increased, alleviating the severe imbalance observed in the original dataset. In contrast, the proportions of necrotic lesions and background remain unaffected, since augmented patches are pasted exclusively onto healthy leaf regions. This ensures that additional samples are introduced primarily for the rare insect-damage class without disrupting the overall composition of other categories. Figure~\ref{fig:rpcp_effect}(b) shows the t-SNE visualization of RGB pixels across different classes. Specifically, the augmented patches are generated by extracting insect-damage regions, pasting them onto healthy leaves, and applying random projection for visual consistency. This observation indicates that the augmented samples are visually plausible and remain within the natural data distribution, supporting their reliability when incorporated as additional training samples.

\section{Conclusion}
In this paper, we have proposed Random Projected Copy-and-Paste (RPCP), a rare-class–oriented augmentation framework for wheat leaf disease segmentation. 
By combining category-aware copy-paste with random-projection refinement, RPCP generates realistic augmented images that enhance the representation of rare classes.
Extensive experiments across diverse segmentation methods demonstrate that RPCP consistently improves rare-class performance while maintaining accuracy on common classes. These results highlight RPCP as a scalable and model-agnostic augmentation strategy for robust plant pathology segmentation.

\subsubsection{\ackname} This work was partially supported by the Analytics for the Australian Grains Industry (AAGI) Strategic Partnership with funding allocated by the Grains Research Development Corporation (GRDC UOQ2301-010OPX) and by The University of Queensland (DVCR2201A).


\bibliographystyle{splncs04}
\bibliography{refs}

\end{document}